%% file: CFRL.tex
\pdfoutput=1

\documentclass[11pt]{article}

\usepackage{acl}

\usepackage{times}
\usepackage{latexsym}
\usepackage{amsmath}
\usepackage[T1]{fontenc}

\usepackage[utf8]{inputenc}

\usepackage{microtype}
\usepackage{booktabs}
\usepackage{comment}

\usepackage{algorithm,algpseudocode}

\usepackage{enumerate}
\usepackage{enumitem}
\usepackage{multirow}
\usepackage{graphicx}
\usepackage{float}
\usepackage{array}
\usepackage{pdfpages}
\input{math-com}

\usepackage[nameinlink]{cleveref}
\crefformat{section}{\S#2#1#3} 
\crefname{algorithm}{Alg.}{Algs.}
\crefformat{subsection}{\S#2#1#3}
\Crefname{equation}{Eq.}{Eqs.}
\Crefname{figure}{Fig.}{Figs.}

\newcommand{\crfl}{\textsc{cfrl}}
\newcommand{\daer}{\textsc{ERDA}}
\newcommand{\crl}{\textsc{crl}}

\title{Continual Few-shot Relation Learning via Embedding Space Regularization and Data Augmentation}

\author{Chengwei Qin$^\clubsuit$ and Shafiq Joty$^\clubsuit$$^\spadesuit$\\
$^\clubsuit$ Nanyang Technological University\\
$^\spadesuit$ Salesforce Research \\
\texttt{\{chengwei003@e.ntu, srjoty@ntu\}.edu.sg}
}

\begin{document}
\maketitle
\begin{abstract}
Existing continual relation learning (\crl) methods rely on plenty of labeled training data for learning a new task, which can be hard to acquire in real scenario as getting large and representative labeled data is often expensive and time-consuming. It is therefore necessary for the model to learn novel relational patterns with very few labeled data while avoiding catastrophic forgetting of previous task knowledge. In this paper, we formulate this challenging yet practical problem as continual few-shot relation learning (\crfl). Based on the finding that learning for new emerging few-shot tasks often results in feature distributions that are incompatible with previous tasks' learned distributions, we propose a novel method 
based on embedding space regularization and data augmentation. Our method generalizes to new few-shot tasks and avoids catastrophic forgetting of previous tasks by 
enforcing extra constraints on the relational embeddings and by adding extra {relevant} data in a self-supervised manner.
With extensive experiments we demonstrate that our method can significantly outperform previous state-of-the-art methods in \crfl\ task settings.\footnote{Code and models are available at {\small \href{https://github.com/qcwthu/Continual_Fewshot_Relation_Learning}{https://github.com/qcwthu/Continual\_Fewshot\_Relation\_Learning}}}
\end{abstract}
\input{Sections/Intro}
\input{Sections/Rel-work}
\input{Sections/Method}
\input{Sections/exp}

\section{Conclusion}

We have introduced continual few-shot relation learning (\crfl), a challenging yet practical problem where the model needs to learn new relational knowledge with very few labeled data continually. We have proposed a novel method, named \daer, to alleviate the over-fitting and catastrophic forgetting problems which are the core issues in \crfl.
{{\daer} imposes relational constraints in the embedding space with innovative losses and adds extra informative data for few-shot tasks in a self-supervised manner to better grasp novel relational patterns and remember previous knowledge.} Extensive experimental results and analysis show that \daer\ significantly outperforms previous methods in all \crfl\ settings investigated in this work. In the future, we would like to investigate ways to combine meta-learning with \crfl.

\bibliography{anthology,custom}
\bibliographystyle{acl_natbib}



\appendix
\input{Sections/appendix}
\end{document}

%% file: math-com.tex
\usepackage{amsmath,amsfonts,bm}

\newcommand{\ie}{{\em i.e.,}}
\newcommand{\eg}{{\em e.g.,}}

\newcommand{\Ni}{({\em i})~}
\newcommand{\Nii}{({\em ii})~}
\newcommand{\Niii}{({\em iii})~}

\newcommand{\Na}{({\em a})~}
\newcommand{\Nb}{({\em b})~}
\newcommand{\Nc}{({\em c})~}
\newcommand{\Nd}{({\em d})~}
\newcommand{\Ne}{({\em e})~}
\newcommand{\Nf}{({\em f})~}
\newcommand{\Ng}{({\em g})~}

\newcommand{\red}[1]{\textcolor{purple}{#1}}
\newcommand{\blue}[1]{\textcolor{blue}{#1}}

\makeatletter   
\newcommand{\sveryshortarrow}[1][3pt]{\mathrel{%
    \vcenter{\hbox{\rule[-.5\fontdimen8\scriptfont3]
               {\scriptratio\dimexpr#1\relax}{\fontdimen8\scriptfont3}}}%
   \mkern-4mu\hbox{\let\f@size\sf@size\usefont{U}{lasy}{m}{n}\symbol{41}}}}
\makeatother

\def\eqref#1{equation~\ref{#1}}

\def\1{\bm{1}}

\def\m1{{\bm{1}}}

\DeclareMathAlphabet{\mathsfit}{\encodingdefault}{\sfdefault}{m}{sl}
\SetMathAlphabet{\mathsfit}{bold}{\encodingdefault}{\sfdefault}{bx}{n}

\def\gT{{\mathcal{T}}}

\def\gV{{\mathcal{V}}}

\def\sT{{\mathbb{T}}}

\DeclareMathOperator{\real}{\rm I\!R}


%% file: Sections/Intro.tex
\section{Introduction}

\textbf{Relation Extraction} (RE) aims to detect the relationship between two entities in a sentence, for example, predicting the relation \emph{birthdate} in the sentence  
``\emph{Kamala Harris} was born in Oakland, California, on \emph{October 20, 1964}.'' for the two entities \emph{Kamala Harris} and \emph{October 20, 1964}. It serves as a fundamental step for  downstream tasks such as search and question answering \citep{dong-etal-2015-question,yu-etal-2017-improved}. Traditionally, RE methods were built by considering a fixed static set of relations \citep{miwa-bansal-2016-end,han-etal-2018-hierarchical}. However, similar to entity recognition, RE is also an open-vocabulary problem \citep{sennrich-etal-2016-neural}, {where the relation set keeps growing as new relation types emerge with new data.}

\begin{figure}[t]
    \centering
    \includegraphics[width=0.46\textwidth]{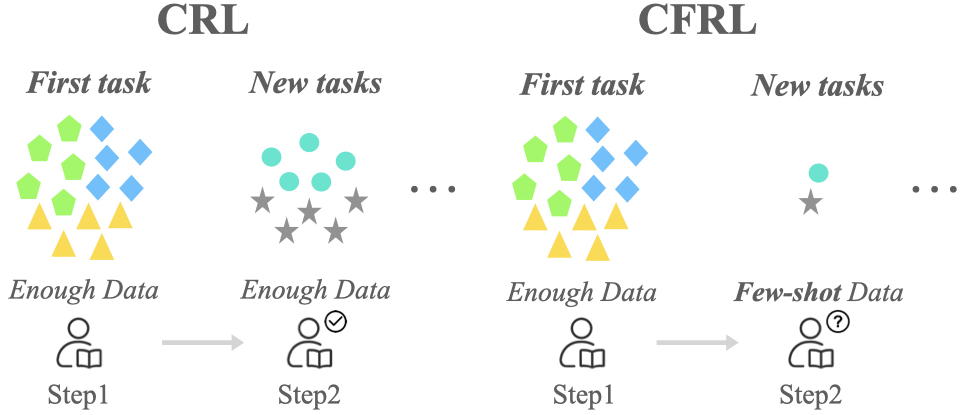}
    \caption{Difference between Continual Relation Learning (\crl) and Continual Few-shot Relation Learning (\crfl). Except for the first task which has enough training data, the subsequent new tasks are all \textit{few-shot} in {\crfl}. In contrast, {\crl} assumes enough training data for every task.} 
\label{fig:comparision}
\end{figure}

A potential solution is to formalize RE as {Continual Relation Learning} or {\crl} \citep{wang2019sentence}. In \crl, the model learns relational knowledge through a sequence of tasks, where the relation set changes dynamically from the current task to the next. The model is expected to perform well on both the novel and previous tasks, which is challenging due to the existence of \textit{Catastrophic Forgetting} phenomenon \citep{mccloskey1989catastrophic,french1999catastrophic} in continual learning. In this phenomenon, the model forgets previous relational knowledge after learning new relational patterns.

Existing methods to address catastrophic forgetting in \crl\ can be divided into three categories: \Ni regularization-based methods, \Nii architecture-based methods, and \Niii memory-based methods. Recent work shows that memory-based methods which save several key examples from previous tasks to a memory and reuse them when learning new tasks are more effective in NLP \citep{wang2019sentence,sun2019lamol}. Successful memory-based \crl\ methods include EAEMR \citep{wang2019sentence}, MLLRE \citep{obamuyide-vlachos-2019-meta}, EMAR \citep{han-etal-2020-continual}, and CML \citep{wu2021curriculum}. 

Despite their effectiveness, one major limitation of these methods is that they all assume plenty of training data for learning new relations (tasks), which is hard to satisfy in real scenario {where continual learning is desirable,} as acquiring large labeled datasets for every new relation is expensive and sometimes impractical for quick deployment (\eg\ RE from news articles during the onset of an emerging event like Covid-19). {In fact, one of the main objectives of continual learning is to {quickly adapt} to new environments or tasks by exploiting previously acquired knowledge, a hallmark of human intelligence} \cite{lopez2017gradient}. If the new tasks are \emph{few-shot}, the existing methods suffer from {over-fitting} as shown later in our experiments (\cref{sec:exp}). Considering that humans can acquire new knowledge from a handful of examples, it is  expected for the models to generalize well on the new tasks with few data. We regard this problem as {Continual Few-shot Relation Learning} or {\crfl} (\red{\Cref{fig:comparision}}). {Indeed, in relation to \crfl,} \newcite{zhang2021few}, \newcite{zhu2021self} and \newcite{chen2021incremental} recently introduce methods for incremental few-shot learning in Computer Vision.
 
Based on the observation that the learning of emerging few-shot tasks may result in distorted feature distributions of new data which are incompatible with previous embedding space \citep{ren-etal-2020-two}, this work introduces a novel model based on {Embedding space Regularization and Data Augmentation} ({\daer}) for \crfl. In particular, we propose a multi-margin loss and a pairwise margin loss in addition to the cross-entropy loss to impose further relational constraints in the embedding space. We  also introduce a novel contrastive loss to learn more effectively from the memory data. Our proposed data augmentation method selects relevant samples from unlabeled text to provide more relational knowledge for the few-shot tasks. The empirical results show that our method can significantly outperform previous state-of-the-art methods. In summary, our main contributions are: 
\begin{itemize}[leftmargin=*,topsep=2pt,itemsep=2pt,parsep=0pt]
    \item To the best of our knowledge, we are the first one to consider \crfl. We define the \crfl\ problem and construct a benchmark for the problem. 
    \item We propose \daer, a novel method for \crfl\ based on 
    {embedding space regularization and data augmentation.}
    \item With extensive experiments, we demonstrate the effectiveness of our method compared to existing ones and analyse our results thoroughly.
\end{itemize}

%% file: Sections/Rel-work.tex
\section{Related Work}

Conventional RE methods include supervised  \citep{zelenko-etal-2002-kernel,liu2013convolution,zeng-etal-2014-relation,
miwa-bansal-2016-end}, 
semi-supervised \citep{chen-etal-2006-relation,
sun-etal-2011-semi,hu2020semi} and distantly supervised methods \citep{mintz-etal-2009-distant,
yao-etal-2011-structured,zeng-etal-2015-distant,han-etal-2018-hierarchical}. 
These methods rely on a predefined relation set and have limitations in real scenario where novel relations are emerging. There have been some efforts which focus on relation learning without predefined types, including open RE \citep{shinyama-sekine-2006-preemptive,etzioni2008open,
cui-etal-2018-neural,gao2020neural} and continual relation learning \citep{wang2019sentence,obamuyide-vlachos-2019-meta,han-etal-2020-continual,wu2021curriculum}.

\noindent\textbf{Continual Learning} (CL) aims to learn knowledge from a sequence of tasks. The main problem CL attempts to address is \emph{catastrophic forgetting} \citep{mccloskey1989catastrophic}, \ie\ the model forgets previous knowledge after learning new tasks. 
{Prior} methods to alleviate this problem can be {mainly} divided into three categories. First, \emph{regularization-based} methods impose constraints on the update of neural weights important to previous tasks to alleviate catastrophic forgetting \citep{li2017learning,kirkpatrick2017overcoming,zenke2017continual,ritter2018online}. Second, \emph{architecture-based} methods  dynamically change model architectures to acquire new information while remembering previous knowledge \citep{chen2015net2net,rusu2016progressive,fernando2017pathnet,mallya2018piggyback}. Finally,  \emph{memory-based} methods  maintain a memory to save key samples of previous tasks to prevent forgetting \citep{rebuffi2017icarl,lopez2017gradient,shin2017continual,chaudhry2018efficient}. 
{These methods mainly focus on learning a single type of tasks. More recently, researchers have considered lifelong language learning \citep{sun2019lamol,qin2022lfpt}, where the model is expected to continually learn from different types of tasks.} 


\noindent\textbf{Few-shot Learning} (FSL) aims to solve tasks containing only a few labeled samples, which faces the issue of over-fitting. To address this, existing methods have explored three different directions: \Ni \emph{data-based} methods use prior knowledge to augment data to the few-shot set \citep{santoro2016meta,
benaim2018one,gao2020neural}; \Nii \emph{model-based} methods reduce the hypothesis space using prior knowledge \citep{rezende2016one,triantafillou2017few,hu-etal-2018-shot}; and \Niii \emph{algorithm-based} methods try to find a more suitable strategy to search for the best  hypothesis in the whole hypothesis space \citep{hoffman2013one,
ravi2016optimization,finn2017model}.

\noindent\textbf{Summary.} {Existing work in \crl\ which involves a sequence of tasks containing \emph{sufficient} training data, mainly focuses on alleviating the catastrophic forgetting of previous relational knowledge when the model is trained on new tasks. The work in few-shot learning mostly leverages prior knowledge to address the over-fitting of novel few-shot tasks. In contrast to these lines of work, we aim to solve a more challenging yet more practical problem \crfl\ where the model needs to learn relational patterns from a sequence of few-shot tasks continually}.

%% file: Sections/Method.tex
\section{Methodology}

In this section, we first formally define the \crfl\ problem. Then, we present our method for  \crfl. 

\subsection{Problem Definition}

\crfl\ involves learning from a sequence of tasks $\sT = (\gT^1,  \ldots, \gT^n)$, where every task $\mathcal{T}^k$ has its own training set $D_{\text{train}}^k$, validation set $D_{\text{valid}}^k$, and test set $D_{\text{test}}^k$.  Each dataset $D$ contains several samples $\{ (x_i,y_i) \}_{i=1}^{|D|}$, whose labels $y_i$ belong to the relation set $R^k$ of task $\mathcal{T}^k$. In contrast to the previously addressed continual relation learning (\crl), \crfl\ assumes that except for the first task which has enough data for training, the subsequent new tasks are all \emph{few-shot}, meaning that they have only few labeled instances ({see \Cref{fig:comparision}}). For example, consider there are three relation learning tasks $\mathcal{T}^1,\mathcal{T}^2$ and $\mathcal{T}^3$ with their corresponding relation sets $R^1,R^2$, and $R^3$, each having $10$ relations. In \crfl, we assume the existing task $\mathcal{T}^1$ has enough training data (\eg\ $100$ samples for every relation in $R^1$), while the new tasks $\mathcal{T}^2$ and $\mathcal{T}^3$ are few-shot with only few (\eg\ $5$) samples for every relation in $R^2$ and $R^3$. Assuming that the relation number of each few-shot task is $N$ and the sample number of every relation is $K$, we call this  setup \textbf{$N$-way $K$-shot} continual learning. The problem setup of \crfl\ is aligned with the real scenario, where we generally have sufficient data for an existing task, but only few labeled data as new tasks emerge.

The model in \crfl\ is expected to first learn $\mathcal{T}^1$ well, which has sufficient training data to obtain good ability to extract the relation information in the sentence. Then at time step $k$, the model will be trained on the training set $D_{\text{train}}^k$ of few-shot task $\mathcal{T}^k$. After learning $\mathcal{T}^k$,  the model is expected to perform well on both $\mathcal{T}^k$ and the previous $k$$-$$1$ tasks, as the model will be evaluated on $\hat D_{\text{test}}^k = \cup_{i=1}^{k} D_{\text{test}}^i$ consisting of all known relations after learning $\mathcal{T}^k$, \ie\ $\hat R^k = \cup_{i=1}^{k} R^i$.  This requires the model to overcome the \emph{catastrophic forgetting} of previous knowledge and to learn new knowledge well with very few labeled data. 

To overcome the catastrophic forgetting problem, a memory $\mathcal{M} = \left\{\mathcal{M}^1,\mathcal{M}^2,...\right\}$, which stores some key samples of previous tasks is maintained during the learning. When the model is learning $\mathcal{T}^k$, it has access to the data saved in memory $\mathcal{M}^1,...,\mathcal{M}^{k-1}$. As there is no limit on the number of tasks, the size of memory $\mathcal{M}^k$ is constrained to be small. Therefore, the model has to select only key samples from the training set $D_{\text{train}}^k$ to save them in $\mathcal{M}^k$.  In our \crfl\ setting, only one sample per relation is allowed to be saved in the memory.

\begin{figure}[t]
    \centering
    \includegraphics[width=0.46\textwidth]{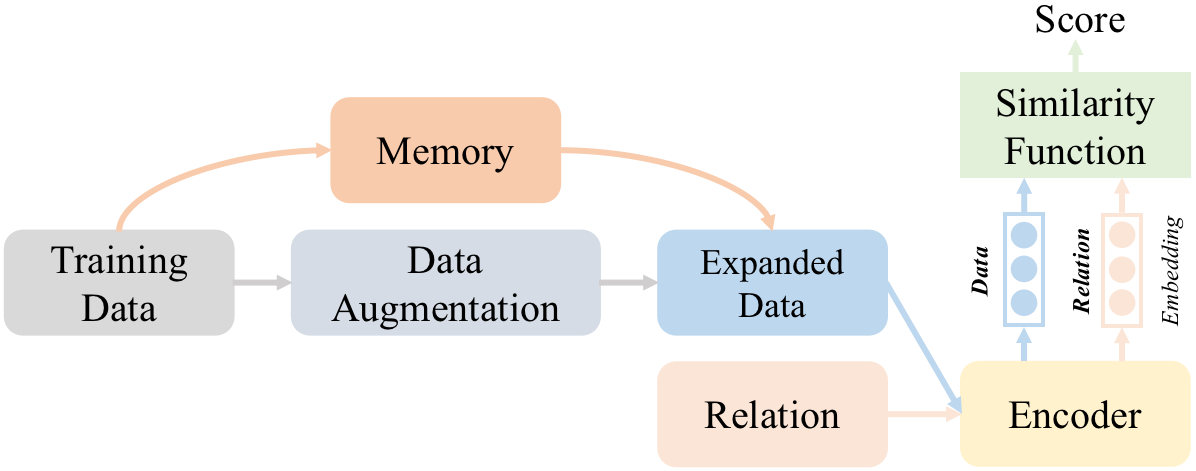}
    \label{Fig.model_arch}
    \caption{\label{model_arch}
    \small{Our framework for \crfl. The Data Augmentation component is used only for few-shot tasks ($k>1$).
    }
    }
\end{figure}

\subsection{Overall Framework}

Our framework for \crfl\ is shown in \Cref{model_arch} and \Cref{alg:Framwork} describes the overall training process (see Appendix ~\ref{sec:appendix2} for a block diagram). At time step $k$, given the training data $D_{\text{train}}^k$ for the task $\mathcal{T}^k$, depending on whether the task is a few-shot or not, the process has four or three working modules, respectively. The general learning process (\cref{sec:gen}) has three steps that apply to all tasks. If the task is a few-shot task ($k>1$), we apply an additional step to create an augmented training 
set $\widetilde D_{\text{train}}^k$. For the initial task ($k=1$), we have $\widetilde D_{\text{train}}^k = D_{\text{train}}^k$.

For any task $\mathcal{T}^k$, we use a siamese model to encode every new relation $r_i \in R^k$ into $\mathbf{r}_i \in \real^{d}$ as well as the sentences, and train the model on $\widetilde D_{\text{train}}^k$ to acquire relation information of the new data
(\cref{section:3.3.2}). To overcome forgetting, we select the most informative sample for each relation $r_i \in R^k$ from $D_{\text{train}}^k$ and update the memory $\hat {\mathcal{M}^k}$ (\cref{section:3.3.3}). Finally, we combine $\widetilde D_{\text{train}}^k$ and $\hat {\mathcal{M}^k}$ as the training data for learning new relational patterns and remembering previous knowledge (\cref{section:3.3.4}). {We also simultaneously update the representation of all relations in 
$\hat R^{k}$, which involves making a forward pass through the current model. The learning and updating are done iteratively for convergence.}

For data augmentation in few-shot tasks (\cref{subsec:da}), we select reliable samples with high relational similarity score from an unlabelled Wikipedia corpus 
using a fine-tuned BERT \citep{devlin-etal-2019-bert}, which serves as the relational similarity model $\mathcal{S}_{\pi}$.  In the interests of coherence, we first present the general learning method followed by the augmentation process for few-shot learning. 

\begin{algorithm}[t!]
\caption{\small Training process at time step $k$}
\small 
\begin{algorithmic}[1] 
    \Require the training set $D_{\text{train}}^k$ and the relation set $R^k$ of the current task $\mathcal{T}^k$, the current memory $\hat {\mathcal{M}}^{k-1}$ and the known relation set $\hat R^{k-1}$, the model $\theta$, the similarity model $\mathcal{S}_{\pi}$, and the unlabeled text corpus $\mathcal{C}$.
    \If{$k == 1$} \Comment{\blue{initial task}}
        \State $\widetilde D_{\text{train}}^k = D_{\text{train}}^k$
    \Else \Comment{\blue{few-shot task}}
        \State \textsc{Select} similar samples from $\mathcal{C}$ using $\mathcal{S}_{\pi}$ for every sample in $D_{\text{train}}^k$ and store them in $A$
        \State $\widetilde D_{\text{train}}^k = A \cup D_{\text{train}}^k$
    \EndIf
    \State {\textsc{Initialize} $\mathbf{r}_i$ for every relation $r_i \in R^k$}
    \For{$i = 1, \ldots, {iter}_1$}
        \State \textsc{Update} $\theta$ with $\mathcal{L}_{\text{new}}$ on $\widetilde D_{\text{train}}^k$ \Comment{\blue{Train on new task}}
    \EndFor
    \State \textsc{Select} key samples from $D_{\text{train}}^k$ for every relation $r_i \in R^k$  to save in $\mathcal{M}^k$
    \State $\hat R^{k} = \hat R^{k-1} \cup R^k$ 
    \State $\hat {\mathcal{M}^{k}} = \hat {\mathcal{M}}^{k-1} \cup \mathcal{M}^k$ \Comment{\blue{Update memory}}
    \State  $\widetilde H^{k} = \widetilde D_{\text{train}}^k \cup \hat {\mathcal{M}^{k}}$ \Comment{\blue{Combine two data sources}}
	\For{$i = 1, \ldots, {iter}_2$}
		\State \textsc{Update} $\theta$ with $\mathcal{L}_{\text{mem}}$ on $\widetilde H^{k}$
		\State \textsc{Update} $\mathbf{r}_i$ for every relation $r_i  \in \hat R^{k}$
	\EndFor
\end{algorithmic}
\label{alg:Framwork} 
\normalsize
\end{algorithm}

\subsection{General Learning Process} \label{sec:gen}

We first introduce the encoder network as it is the {basic} component of the whole framework. 

\subsubsection{The Encoder Network} \label{section:3.3.1}

The {siamese} encoder ($f_{\theta}$) aims at extracting generic and relation related features from the input. The input can be a labeled sentence or the name of a relation. We adopt two kinds of encoders:

\paragraph{ $\bullet$ Bi-LSTM} To have a fair comparison with previous work, we use the same architecture as \newcite{han-etal-2020-continual}. It takes GloVe embeddings \citep{pennington-etal-2014-glove} of the words in a given input and produces a vector representation through a Bi-LSTM \citep{hochreiter1997long}.

\paragraph{$\bullet$ BERT} 

We adopt $\rm{BERT_{base}}$ which has 12 layers and 110M parameters. As the new tasks are few-shot, we only fine-tune the 12-th encoding layer and the extra linear layer. We include special tokens  
around the entities (`\#' for the head entity and `@' for the tail entity) in a given labeled sentence to improve the encoder's understanding of relation information. We use the {\sc{[cls]}} token features as the representation of the input sequence. 

\subsubsection{Learning with New Data} \label{section:3.3.2}

At time step $k$, to have a good understanding of the new relations, we fine-tune the model on the expanded dataset $\widetilde D_{\text{train}}^{k}$. The model $f_{\theta}$ first encodes the name of each new relation $r_j \in R^k$ into its representation $\mathbf{r}_j \in \real^{d}$ by making a forward pass. Then, we optimize the parameters ($\theta$) by minimizing a loss $\mathcal{L}_{\text{new}}$ that consists of a cross entropy loss, a multi-margin loss and a pairwise margin loss.

The \textbf{cross entropy} loss $\mathcal{L}_{\text{ce}}$ is used for relation classification as follows.
\begin{equation}
\small 
\begin{aligned}
 \sum_{(x_i,y_i) \in \widetilde D_{\text{train}}^{k}}   \sum_{j=1}^{| \hat R^k |} \delta_{y_i,r_j}  \times       \log \frac{\exp (g(f_{\theta}(x_i),\mathbf{r}_j))}{\sum_{l=1}^{| \hat R^k |}\exp (g(f_{\theta}(x_i),\mathbf{r}_l))} \label{eq:ce}
\end{aligned}
\normalsize
\end{equation}
\noindent where $\hat R^k$ is the set of all known relations at step $k$, $g(,)$ is a function used to measure similarity between two vectors (\eg \ cosine similarity or L2 distance), and $\delta_{a,b}$ is the Kronecker delta function-- $\delta_{a,b} = 1$ if $a$ equals $b$, otherwise $\delta_{a,b} = 0$.

In inference, we choose the relation label that has the highest similarity with the input sentence (Eq. \ref{eq:inf}). To ensure that an example has the highest similarity with the true relation, we additionally design two \textbf{margin-based} losses, which increase the score between an example and the true label while decreasing the scores for the wrong labels. The first one is a \textbf{multi-margin} loss defined as:
\begin{equation}
\small 
\begin{aligned}
   \mathcal{L}_{\text{mm}} & =  \sum_{(x_i,y_i) \in \widetilde D_{\text{train}}^{k}} \sum_{j=1,j \neq t_i}^{| \hat R^k |} \max \Big( 0,  \\
    &  m_1 - g(f_{\theta}(x_i),\mathbf{r}_{t_i}) + g(f_{\theta}(x_i),\mathbf{r}_j) \Big)
\end{aligned}
\normalsize
\end{equation}
where $t_i$ is the correct relation index in $\hat R^k$ satisfying $r_{t_i} = y_i$ and $m_1$ is a margin value. The $\mathcal{L}_{\text{mm}}$ loss attempts to ensure intra-class compactness while increasing inter-class distances. The second one is a \textbf{pairwise margin} loss $\mathcal{L}_{\text{pm}}$:
\begin{equation}
\small 
\begin{aligned}
 \sum_{~~(x_i,y_i) \in \widetilde D_{\text{train}}^{k}}  \max \Big( 0,  m_2 - g(f_{\theta}(x_i),\mathbf{r}_{t_i}) + g(f_{\theta}(x_i),\mathbf{r}_{s_i}) \Big)
\end{aligned}
\end{equation}
where $m_2$ is the margin for $\mathcal{L}_{\text{pm}}$ and $s_i = \mathop{\arg\max}_{s} g(f_{\theta}(x_i),\mathbf{r}_{s})$ s.t. $s \neq t_i$,  the closest wrong label. The $\mathcal{L}_{\text{pm}}$ loss penalizes the cases where the similarity score of the closest wrong label is higher than the score of the correct label \citep{yang2018robust}. Both $\mathcal{L}_{\text{mm}}$ and $\mathcal{L}_{\text{pm}}$ improve the discriminative ability of the model (\cref{subsec:abl}).

\noindent The \textbf{total loss} for learning on $\mathcal{T}^k$ is defined as:
\begin{equation}
\small
    \mathcal{L}_{new} = \lambda_{\text{ce}} \mathcal{L}_{\text{ce}} + \lambda_{\text{mm}} \mathcal{L}_{\text{mm}} + \lambda_{\text{pm}} \mathcal{L}_{\text{pm}} \label{Lnew}
\normalsize    
\end{equation}
where $\lambda_{\text{ce}}$, $\lambda_{\text{mm}}$ and $\lambda_{\text{pm}}$ are the relative weights of the component losses, respectively.

\subsubsection{Selecting Samples for Memory} \label{section:3.3.3}

After training the model $f_{\theta}$ with Eq. (\ref{Lnew}), we use it to select one sample per new relation. Specifically, for every new relation $r_j \in R^k$, we obtain the centroid feature $\mathbf{c}_j$ by averaging the embeddings of all samples labeled as $r_j$ in $D_{\text{train}}^{k}$ as follows.
\begin{equation}
\small 
    \mathbf{c}_j =  \frac{1}{|D_{r_j}^k|}\sum_{(x_i,y_i) \in D_{r_j}^k} f_{\theta}(x_i)
\normalsize
\end{equation}
where $D_{r_j}^k = \{(x_i,y_i)|(x_i,y_i) \in D_{\text{train}}^{k}, y_i = r_j\}$. Then we select the instance closest to $\mathbf{c}_j$ from $D_{r_j}^k$ as the most informative sample and save it in memory $\mathcal{M}^k$. Note that the selection is done from $D_{\text{train}}^{k}$, not from the expanded set $\widetilde D_{\text{train}}^{k}$.

\subsubsection{Alleviating Forgetting through Memory} \label{section:3.3.4}

As the learning of new relational patterns may cause catastrophic forgetting of previous knowledge (see baselines in \cref{sec:exp}), our model needs to learn 
from the memory data to alleviate forgetting. We combine the expanded set $\widetilde D_{\text{train}}^{k}$ and the whole memory data $\hat {\mathcal{M}^k} = \cup_{j=1}^{k} \mathcal{M}^j$ into $\widetilde H^{k}$ to allow the model to learn new relational knowledge and consolidate previous knowledge. However, the memory data is limited containing only one sample per relation. To learn effectively from such limited data, we design a novel method  to generate a \emph{hard negative sample} set ${P}_i$ for every sample in $\hat {\mathcal{M}^k}$.

The negative samples are generated on the fly. After sampling a mini-batch $B_t$ from $\widetilde H^{k}$, we consider all memory data in $B_t$ as $M_{B_t}$. For every sample $(\hat x_i,\hat y_i)$ in $M_{B_t}$, we replace its head entity $e_i^h$ or tail entity $e_i^t$ with the corresponding entity of a randomly selected sample in the same batch $B_t$ to get the {hard negative} sample set ${P}_i=\{(\hat x_j^{P_i},\hat y_i)\}_{j=1}^{|{P}_i|}$. Then $(\hat x_i,\hat y_i)$ and ${P}_i$ are used to calculate a margin-based \textbf{contrastive loss} $\mathcal{L}_{\text{con}}$ as follows.
\begin{equation}
\small 
\begin{aligned}
    \mathcal{L}_{\text{con}} = \sum_{(\hat x_i,\hat y_i) \in M_{B_t}} & \max \Big( 0, m_3 - g(f_{\theta}(\hat x_i),\mathbf{r}_{\hat t_i}) + \\[-0.8em]
     &  \sum_{(\hat x_j^{P_i},\hat y_i) \in {P}_i}g(f_{\theta}(\hat x_j^{P_i}),\mathbf{r}_{\hat t_i}) \Big)
\end{aligned}
\normalsize
\end{equation}
where $\hat t_i$ is the relation index satisfying $r_{\hat t_i} = \hat y_i$ and $m_3$ is the margin value for $\mathcal{L}_{\text{con}}$. This loss forces the model to distinguish the valid relations from the hard negatives so that the model learns more precise and fine-grained relational knowledge. In addition, we also use the three losses $\mathcal{L}_{\text{ce}}$ and $\mathcal{L}_{\text{mm}}$ and $\mathcal{L}_{\text{pm}}$ defined in \Cref{section:3.3.2} to update $\theta$ on $B_t$. The total loss on the memory data is:
\begin{equation}
\small 
\begin{aligned}
    \mathcal{L}_{\text{mem}}  &= \lambda_{\text{ce}} \mathcal{L}_{\text{ce}} + \lambda_{\text{mm}} \mathcal{L}_{\text{mm}} + \lambda_{\text{pm}} \mathcal{L}_{\text{pm}} + \lambda_{\text{con}} \mathcal{L}_{\text{con}} \label{Lmem}
\end{aligned}
\normalsize
\end{equation}
where $\lambda_{\text{ce}}$, $\lambda_{\text{mm}}$, $\lambda_{\text{pm}}$ and $\lambda_{\text{con}}$ are the relative weights of the corresponding losses.

\paragraph{Updating Relation Embeddings}

{After training the model  on $\widetilde H^{k}$ for few steps, we use the memory $\hat {\mathcal{M}^k}$ to update the relation embedding $\mathbf{r}_i$ of all known relations. For a relation $r_i \in \hat R^k$, we average the embeddings (obtained by making a forward pass through $f_{\theta}$) of the relation name and memory data to obtain its updated representation $\mathbf{r}_i$.} The training of $\theta$ and updating of $\mathbf{r}_i$ is done iteratively to grasp new relational patterns while alleviating the catastrophic forgetting of previous knowledge.

\subsubsection{Inference} \label{section:3.3.5}

For a given input $x_i$ in $\hat D_{\text{test}}^k$, we calculate the similarity between $x_i$ and all known relations, and pick the one with the highest similarity score:
\begin{equation}
\small
    y_i^{*} = \mathop{\arg\max}\limits_{r \in \hat R^k} g(f_{\theta}(x_i),\mathbf{r}) \label{eq:inf}
\normalsize
\end{equation}

\subsection{Data Augmentation for Few-shot Tasks} \label{subsec:da}

For each few-shot task $\mathcal{T}^k$, we aim to get more data by selecting reliable samples from an unlabeled corpus $\mathcal{C}$ with tagged entities before the general learning process  (\cref{sec:gen}) begins. We achieve this using a relational similarity model $\mathcal{S}_{\pi}$ and sentences from Wikipedia as $\mathcal{C}$. The model $\mathcal{S}_{\pi}$ (described later) takes a sentence as input and produces a normalized vector representation. The cosine similarity between two vectors is used to measure the relational similarity between the two corresponding sentences. A higher similarity means the two sentences are more likely to have the same relation label. We propose two novel selection methods, {which are complementary to each other}.

\paragraph{(a) Augmentation via Entity Matching} For each instance $(x_i,y_i)$ in $D_{\text{train}}^k$, we extract its entity pair $(e_i^h,e_i^t)$ with $e_i^h$ being the head entity and $e_i^t$ being the tail entity. As sentences with the same entity pair are more likely to express the same relation, we first collect a candidate set $\mathcal{Q} = \left\{ \widetilde x_j\right\}_{j=1}^{|\mathcal{Q}|}$ from $\mathcal{C}$, where $\widetilde x_j$ shares the same entity pair $(e_i^h,e_i^t)$ with $x_i$. If $\mathcal{Q}$ is a non-empty set, we pair all $\widetilde x_j$ in $\mathcal{Q}$ with $x_i$, and denote each pair as $\langle \widetilde x_j,x_i \rangle$. Then we use $\mathcal{S}_{\pi}$ to obtain a similarity score $\mathrm{s}_j$ for $\langle \widetilde x_j,x_i \rangle$. After getting scores for all pairs, we pick the instances $\widetilde x_j$ with similarity score $\mathrm{s}_j$ higher than a predefined threshold $\alpha$ as new samples and label them with relation $y_i$. The selected instances are then augmented to $D_{\text{train}}^k$ as additional data.

\vspace{-0.5em} 
\paragraph{(b) Augmentation via Similarity Search} 

The hard entity matching could be too restrictive at times. For example, even though the sentences ``\emph{Harry Potter} is written by \emph{Joanne Rowling}'' and ``\emph{Charles Dickens} is the author of \emph{A Tale of Two Cities}'' share the same relation \emph{author}, hard matching fails to find any relevance. Therefore, in cases when entity matching returns an empty $\mathcal{Q}$, we resort to similarity search using Faiss \citep{JDH17}. Given a query vector $\mathbf{q}_i$, it can efficiently search for vectors $\{\mathbf{v}_j\}_{j=1}^{K}$ with the top-$K$ highest similarity scores in a large vector set $\gV$. In our case, $\mathbf{q}_i$ is the representation of $x_i$ and $\mathcal{V}$ contains the representations of the sentences in $\mathcal{C}$. We use $\mathcal{S}_{\pi}$ to obtain these representations; the difference is that $\mathcal{V}$ is pre-computed while $\mathbf{q}_i$ is obtained during training. We labeled the top-$K$ most similar instances with $y_i$ and augment them to $D_{\text{train}}^k$.

\vspace{-0.5em} 
\paragraph{Similarity Model} To train $\mathcal{S}_{\pi}$, inspired by \newcite{soares2019matching}, we adopt a \emph{contrastive learning} method to fine-tune a $\rm{BERT_{base}}$ model on $\mathcal{C}$, whose sentences are already tagged with entities. Based on the observation that sentences with the same entity pair are more likely to encode the same relation, we use sentence pairs containing the same entities in $\mathcal{C}$ as positive samples. For negatives, instead of using all sentence pairs containing different entities, we select pairs sharing only one entity as \textbf{hard negatives} (\ie\ pair $(x_i,x_j)$ where $e_i^h = e_j^h$ and $e_i^t \neq e_j^t$ or $e_i^t = e_j^t$ and $e_i^h \neq e_j^h$ 
). We randomly sample the same number of negative samples as the positive ones to balance the training.

For an input pair $(x_i,x_j)$, we compute the similarity score based on the following formula. 
\begin{equation}
\small 
    \sigma(x_i,x_j) = \frac{1}{1+\exp (-\mathcal{S}_{\pi}(x_i)^T \mathcal{S}_{\pi}(x_j))}
\normalsize
\end{equation}
where $\mathcal{S}_{\pi}(x)$ is the normalized representation of $x$ obtained from the final 
layer of BERT. Then we optimize the parameters $\pi$ of $\mathcal{S}_{\pi}$ by minimizing a binary cross entropy loss $\mathcal{L}_{\text{pretrain}}$ as follows. 
\begin{equation}
\small 
\begin{aligned}
     -  \sum_{(x_i,x_j) \in \mathcal{C}_p}  \log \sigma (x_i,x_j) -  \sum_{(x_i^{\prime},x_j^{\prime}) \in \mathcal{C}_n}  \log (1 - \sigma(x_i^{\prime},x_j^{\prime}))
\end{aligned}
\normalsize
\end{equation}
where $\mathcal{C}_p$ is a positive batch and $\mathcal{C}_n$ is a negative batch. This objective tries to ensure that sentence pairs with the same entity pairs have higher cosine similarity than those with different entities.

%% file: Sections/exp.tex
\section{Experiment} \label{sec:exp}

We define the benchmark and evaluation metric for \crfl\ before presenting our experimental results.

\subsection{Benchmark and Evaluation Metric}

\textbf{Benchmark~~} As the benchmark for \crfl\ needs to have sufficient relations as well as data and be suitable for few-shot learning, we create the \crfl\ benchmark based on \textbf{FewRel} \citep{han-etal-2018-fewrel}. FewRel is a large-scale dataset for few-shot RE, which contains $80$ relations with hundreds of samples per relation. We randomly split the $80$ relations into $8$ tasks, where each task contains $10$ relations (\textit{10-way}). To have enough data for the first task $\mathcal{T}^1$, we sample $100$ samples per relation. All the subsequent tasks $\mathcal{T}^2,...,\mathcal{T}^8$ are few-shot; for each relation, we conduct \textit{2-shot}, \textit{5-shot} and \textit{10-shot} experiments to verify the effectiveness of our method.

In addition, to demonstrate the generalizability of our method, we also create a \crfl\ benchmark based on the \textbf{TACRED} dataset \citep{zhang-etal-2017-position} which contains only $42$ relations. We filter out the special relation ``n/a'' (not available) and split the remaining $41$ relations into $8$ tasks. Except for the first task that contains $6$ relations, all other tasks have $5$ relations (\textit{5-way}). Similar to FewRel, we randomly sample $100$ examples per relation in $\mathcal{T}^1$ and conduct \textit{5-shot} and \textit{10-shot} experiments.

\paragraph{Metric}

At time step $k$, we evaluate the model performance through relation classification accuracy on the test sets $\hat D_{\text{test}}^k = \cup_{i=1}^{k} D_{\text{test}}^i$ of all seen tasks $\{\mathcal{T}^i\}_{i=1}^{k}$. This metric reflects whether the model can alleviate catastrophic forgetting while acquiring novel knowledge well with very few data. Since the model performance might be influenced by task sequences and few-shot training samples, we run every experiment $6$ times each time with a different random seed to ensure a random task order and model initialization, and report the average accuracy along with variance. We perform paired t-test for statistical significance.

\subsection{Model Settings \& Baselines}

The model settings are shown in Appendix \ref{sec:appendix3}.
We compare our approach with the following baselines:

\begin{itemize}[leftmargin=*,topsep=4pt,itemsep=4pt,parsep=0pt]

\item \textbf{SeqRun} fine-tunes the model only on the training data of the new tasks without using any memory data. It may face serious catastrophic forgetting and serves as a \textbf{lower bound}.

\item \textbf{Joint Training} stores all previous samples in the memory and trains the model on all data for each new task. It serves as an \textbf{upper bound} in \textbf{\crl}.

\item \textbf{EMR} \citep{wang2019sentence} maintains a memory for storing selected samples from previous tasks. When training on a novel task, EMR combines the new training data and memory data.

\item \textbf{EMAR} \citep{han-etal-2020-continual} is the state-of-the-art on \crl, which adopts memory activation and reconsolidation to alleviate catastrophic forgetting.

\item \textbf{IDLVQ-C} \citep{chen2021incremental} 
{introduces quantized reference vectors to represent previous knowledge and mitigates catastrophic forgetting by imposing constraints on the quantized vectors and embedded space.} It was originally proposed for image classification with state-of-the-art results in incremental few-shot learning.

\end{itemize}

\begin{table}[t!] \small
\setlength{\tabcolsep}{1.3pt}
\centering
\resizebox{1.00\linewidth}{!}{%
\begin{tabular}{llcccccccc}
\toprule
\multicolumn{2}{l}{\multirow{2}*{\textbf{Method}}}& \multicolumn{8}{c}{\textbf{Task index}}\\
\cmidrule{3-10}
\multicolumn{2}{c}{} & 1 & 2 & 3 & 4 & 5 & 6 & 7 & 8 \\
\midrule
\multicolumn{2}{l}{SeqRun}          & $92.78$ & $52.11$ & $30.08$ & $24.33$ & $19.83$ & $16.90$ & $14.36$ & $12.34$\\
\multicolumn{2}{l}{Joint Train}  & $\mathbf{92.78}$ & $76.29$ & $69.39$ & $64.75$ & $60.45$ & $\mathbf{57.64}$ & $52.80$ & $50.03$\\
\midrule
\multicolumn{2}{l}{EMR}             & $92.78$ & $69.14$ & $56.24$ & $50.03$ & $46.50$ & $43.21$ & $39.88$ & $37.51$ \\
\multicolumn{2}{l}{EMAR}            & $85.20$ & $62.02$ & $52.45$ & $48.95$ & $46.77$ & $44.33$ & $40.75$ & $39.04$\\
\multicolumn{2}{l}{IDLVQ-C}         & $92.23$ & $69.15$ & $57.42$ & $51.66$ & $49.31$ & $46.24$ & $42.25$ & $40.56$ \\
\midrule
\multicolumn{2}{l}{\textbf{\daer}}   & $92.57$ & $\mathbf{79.17}$ & $\mathbf{70.43}$ & $\mathbf{65.01}$ & $\mathbf{61.06}$ & $\mathbf{57.54}$ & $\mathbf{54.88}$ & $\mathbf{53.23}$ \\
\bottomrule
\end{tabular}
}

\caption{
\small Accuracy ($\%$) of different methods at every time step on \textbf{FewRel} benchmark for \textbf{$10$-way $5$-shot} \crfl. \daer\ is significantly better than IDLVQ-C with $p$-value $<0.001$.
}
\label{5shot}
\end{table}

\begin{figure}[tb]
    \centering
    \includegraphics[width=0.98\linewidth]{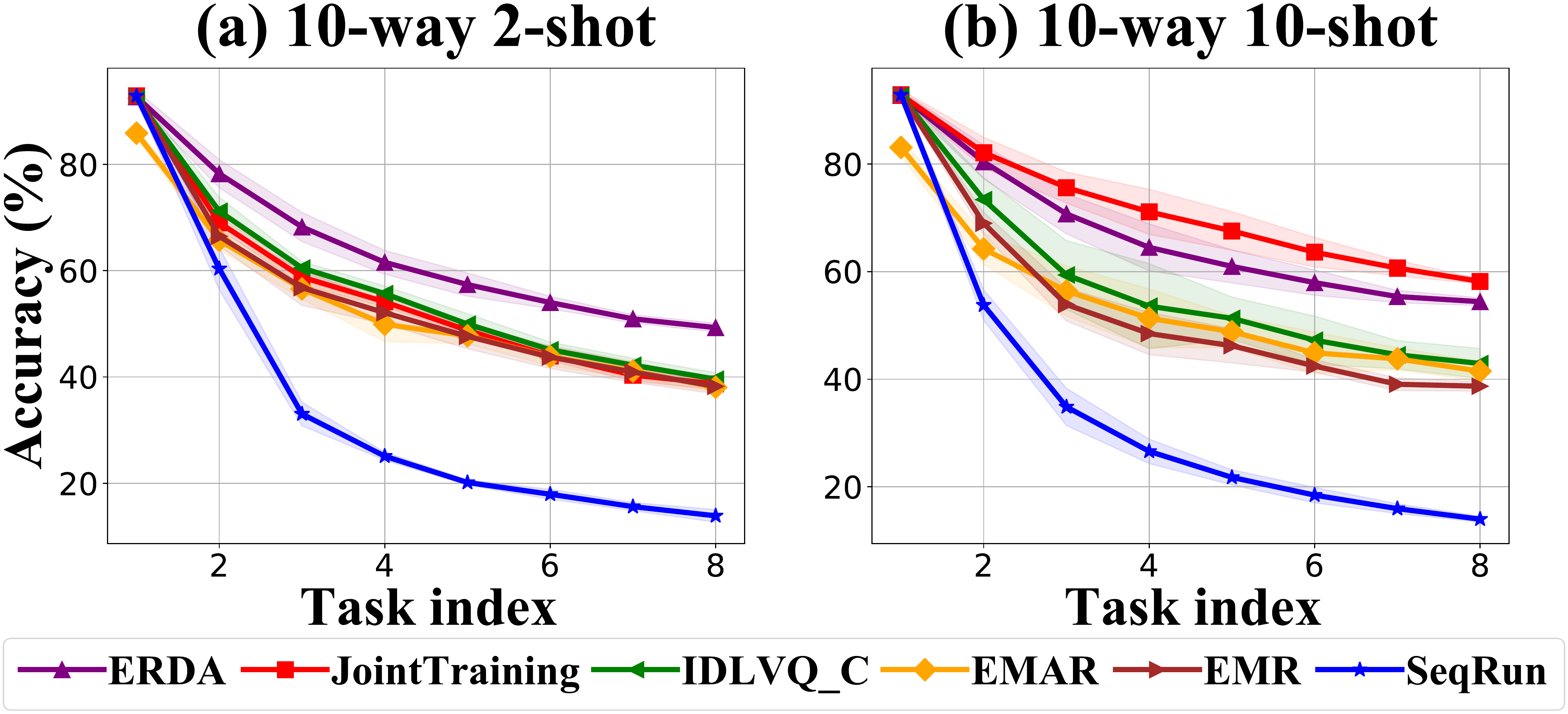}
    \label{Fig.2and10shot}
    \caption{\label{2and10shot}
    \small Comparison results at each time step on \textbf{FewRel} benchmark for \textbf{$10$-way $2$-shot} and \textbf{$10$-shot} settings. For both settings, \daer\ is significantly better than IDLVQ-C with $p$-value $<0.001$. The variance is reported as light color region.
    }
\end{figure}

\subsection{Main Results} \label{main_res}

We compare the performance of different methods using the same setting as EMAR \cite{han-etal-2020-continual}, which uses a Bi-LSTM encoder. We {also} report the results with a BERT encoder.

\paragraph{FewRel Benchmark} We report our results on \textit{10-way 5-shot} in \Cref{5shot}, while \Cref{2and10shot} shows the results on the \textit{10-way} \textit{2-shot} and \textit{10-way} \textit{10-shot} settings.\footnote{To avoid visual clutter, we report only mean scores over 6 runs in \Cref{5shot} and refer to \Cref{5shot-with-var} and \Cref{allruns} in Appendix for variance and elaborate results for different task order.} From the results, we can observe that:

\begin{figure}[t]
    \centering
    \includegraphics[width=0.48\textwidth]{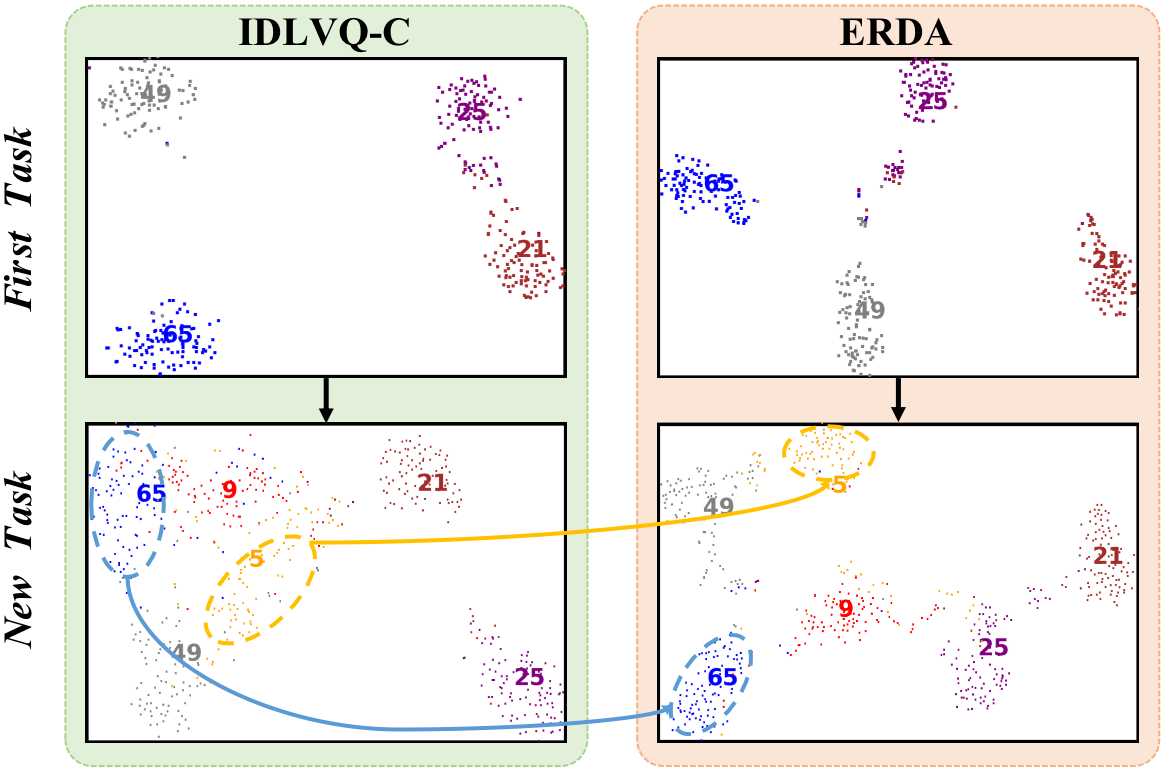}
    \caption{
    \small t-SNE visualization of IDLVQ-C and {\daer} at two stages.  Colors represent  different relation classes with numbers being the relation indices. The initial embeddings of four base classes after learning the first task are shown in the upper row. As the data for the first task is sufficient, both methods can obtain separable embedding space. The lower row shows the embeddings of four base classes and two novel classes (Id 5 and 9) after learning a new few-shot task. Compared with IDLVQ-C, {\daer} shows better intra-class compactness (circled regions) and larger inter-class distances (see the  distances between 5 and 9, and 9 and 65).
    }
    \label{Fig.visualization}
\end{figure}

\noindent $\bullet~$ Our proposed {\daer} outperforms previous baselines in all \crfl\ settings, which demonstrates the superiority of our method. Simply fine-tuning the model with new few-shot examples leads to rapid drops in accuracy due to severe over-fitting and catastrophic forgetting. Although EMR and EMAR adopt a memory module to alleviate forgetting, their performance still decreases quickly as they require plenty of training data for learning a new task. Compared with EMR and EMAR, IDLVQ-C is slightly better as it introduces quantized vectors that can better represent the embedding space of few-shot tasks. {However, IDLVQ-C does not necessarily push the samples from different relations to be far apart in the embedding space and the updating method for the reference vectors may not be optimal. {\daer} outperforms IDLVQ-C by a large margin through embedding space regularization and self-supervised data augmentation.} To verify this, we show the embedding space of IDLVQ-C and {\daer} using t-SNE \citep{van2008visualizing}. We randomly choose four classes from the first task of FewRel and two classes from the new  task, and visualize the test data of these classes  in \Cref{Fig.visualization}. As can be seen, the embedding space obtained by {\daer} shows better intra-class compactness and larger inter-class distances.

\noindent $\bullet~$ Unlike \crl, joint training does not always serve as an upper bound in \crfl\ due to the extremely imbalanced data distribution. Benefiting from the ability to learn feature distribution with very few data, both {\daer} and IDLVQ-C perform better than joint training in the \textit{2-shot} setting. However, as the number of few-shot samples increases, the performance of IDLVQ-C falls far behind joint training, while {\daer} still performs better. In the \textit{5-shot} setting, {\daer} could achieve better results than joint training which verifies the effectiveness of self-supervised data augmentation (more on this in \cref{subsec:abl}). Although {\daer} performs worse than joint training in the \textit{10-shot} setting, its results are still much better than other baselines.

\noindent $\bullet~$ After learning all few-shot tasks, {\daer} outperforms IDLVQ-C by 
{\textbf{9.69}$\bm{\%}$, \textbf{12.67}$\bm{\%}$ and \textbf{11.49}$\bm{\%}$}
in the \textit{2-shot}, \textit{5-shot} and \textit{10-shot} settings, respectively. Moreover, 
the {relative} gain of {\daer} keeps growing with the increasing number of new few-shot tasks. This demonstrates the ability of our method in handling a longer sequence of \crfl\ tasks. 

\begin{figure}[t]
    \centering
    \includegraphics[width=0.98\linewidth]{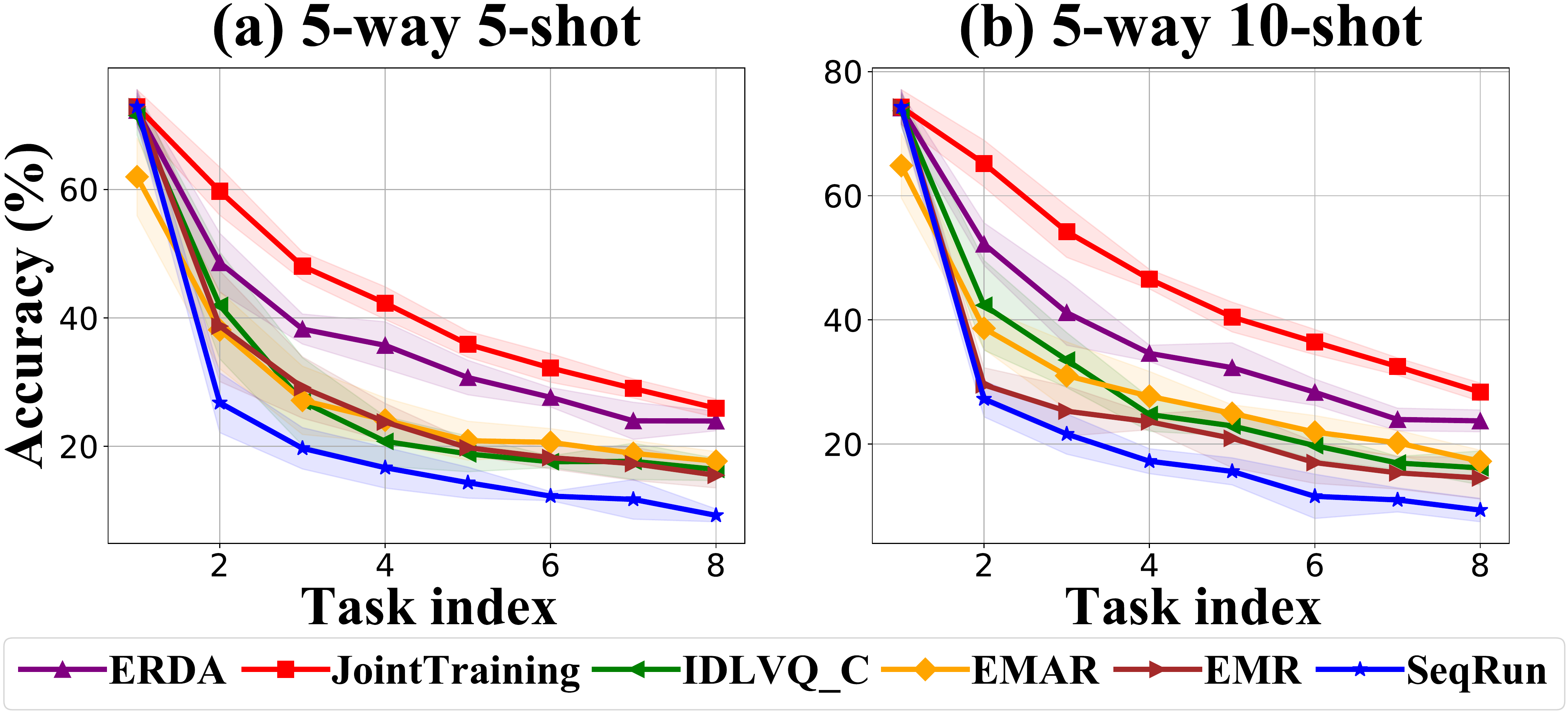}
    \label{Fig.5and_tacred}
    \caption{\label{5and_tacred}
    \small Comparison results at every time step on \textbf{TACRED} benchmark for \textbf{$5$-way $5$-shot} and \textbf{$10$-shot} settings. \daer\ is significantly better than IDLVQ-C with $p$-value $<0.001$ for both settings. The variance is reported as light color region.
    }
\end{figure}

\paragraph{TACRED Benchmark} 

\Cref{5and_tacred} shows the \textit{5-way} \textit{5-shot} and \textit{5-way} \textit{10-shot} results on  TACRED. We can see that  here also {\daer} outperforms all other methods by a large margin which verifies the strong generalization ability of our proposed method.

\begin{figure}[t!]
    \centering
    \includegraphics[width=0.98\linewidth]{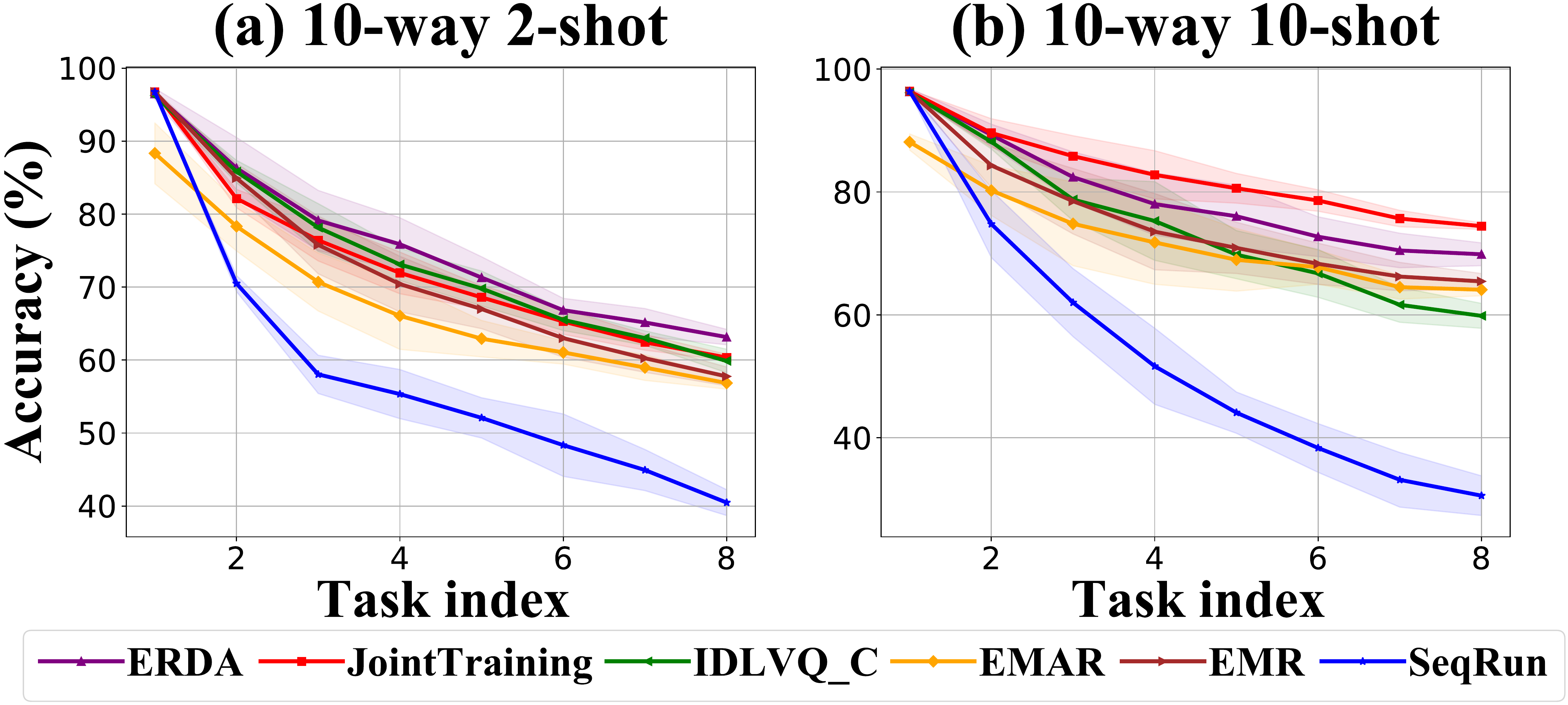}
    \label{Fig.2and10shot}
    \caption{\label{bert-2and10shot}
    \small Comparison results of different methods with a BERT encoder on \textbf{FewRel} benchmark for \textbf{$10$-way $2$-shot} and \textbf{$10$-shot} settings. \daer\ is significantly better than IDLVQ-C with $p$-value $= 0.005$ for $2$-shot setting and is significantly better than EMR with $p$-value $= 0.002$ for $10$-shot setting.
    }
\end{figure}

\begin{figure}[t]
    \centering
    \includegraphics[width=0.98\linewidth]{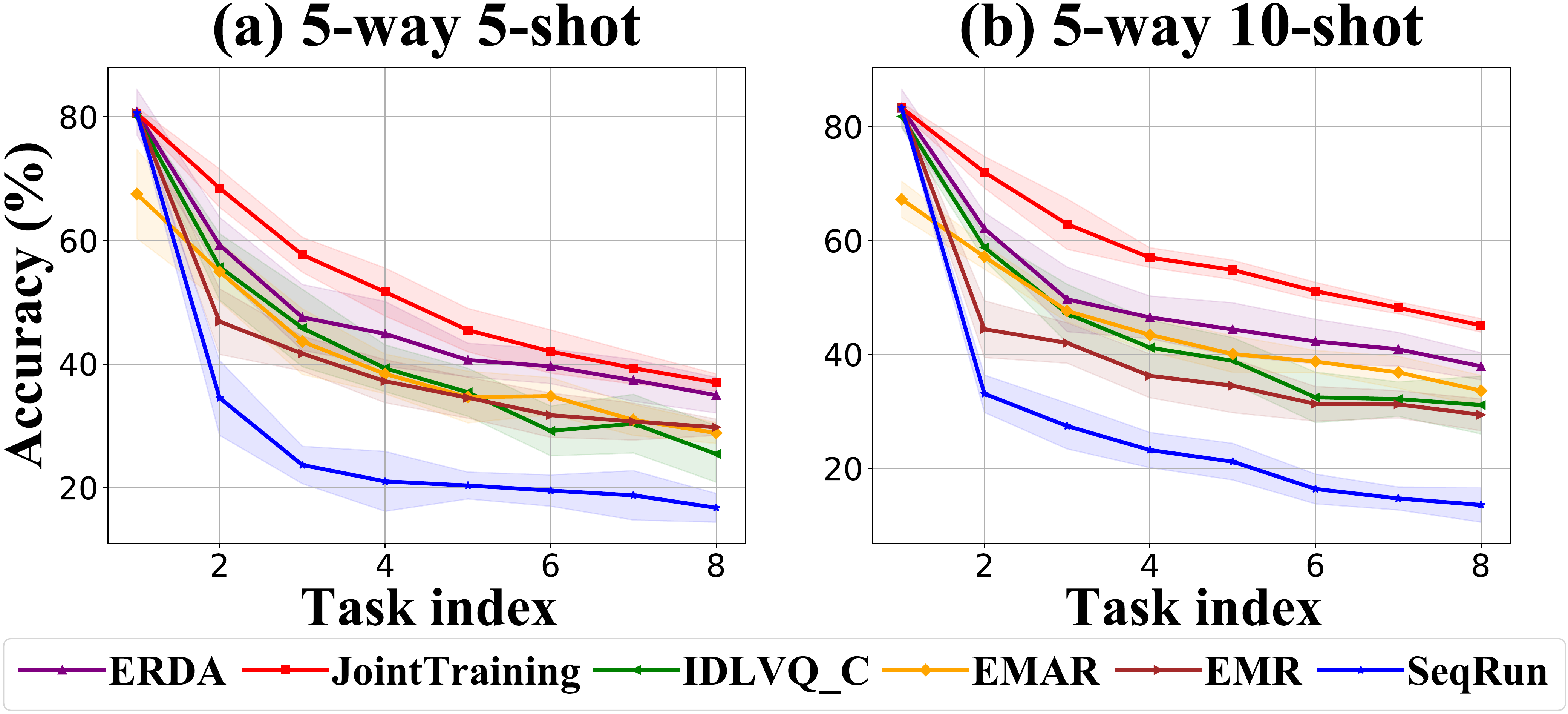}
    \label{Fig.5and_tacred}
    \caption{\label{bert-5and_tacred}
    \small Results of different methods with a BERT encoder on \textbf{TACRED} benchmark for \textbf{$5$-way $5$-shot} and \textbf{$10$-shot} settings. \daer\ is significantly better than EMR with $p$-value $= 0.004$ for $5$-shot setting and is significantly better than EMAR with $p$-value $= 0.02$ for $10$-shot setting.
    }
\end{figure}

\paragraph{Results with BERT} {We show the results with $\rm{BERT_{base}}$ of different methods on FewRel in \Cref{bert-2and10shot} for \textit{10-way} \textit{2-shot} and \textit{10-shot} and \Cref{bert-5shot} for \textit{10-way 5-shot} (in Appendix). The results of on TACRED benchmark are shown in \Cref{bert-5and_tacred} for \textit{5-way} \textit{5-shot} and \textit{10-shot}. From the results, we can observe that ERDA outperforms previous baselines in all \crfl\ settings with a BERT encoder.}

\subsection{Ablation Study} \label{subsec:abl}

\begin{table*}[ht!]
\centering
\resizebox{1.00\linewidth}{!}{
\begin{tabular}{llcccccccc}
\toprule

\multicolumn{2}{l}{$\lambda_{con}$} & 0 & 0.01 & 0.02 & 0.05 & 0.1 & 0.2 & 0.5 & 1.0\\
\midrule
\multicolumn{2}{l}{Accuracy ($\%$)}    & $51.95_{\pm 1.15}$ & $52.66_{\pm 1.23}$ & $\mathbf{53.38}_{\pm 0.63}$ & $53.10_{\pm 0.69}$ & $53.23_{\pm 1.49}$ & $52.99_{\pm 0.79}$ & $52.13_{\pm 1.50}$ & $52.27_{\pm 1.07}$ \\
\bottomrule
\end{tabular}
}
\caption{\label{con_ablation}
\small Accuracy ($\%$) after learning all tasks with different $\lambda_{con}$ on \textbf{FewRel} benchmark (\textbf{$10$-way $5$-shot}).
}
\end{table*}

We conduct several ablations to analyze the contribution of different components of \daer\ on the FewRel \textit{10-way 5-shot} setting. 
In particular, we investigate seven other variants of \daer\ by removing one component at a time:  \Na the multi-margin loss $\mathcal{L}_{\text{mm}}$, \Nb the pairwise margin loss $\mathcal{L}_{\text{pm}}$, \Nc  the margin-based contrastive loss $\mathcal{L}_{\text{con}}$, \Nd the whole 2-stage data augmentation module, \Ne the entity matching method of augmentation, \Nf the similarity search method of augmentation, and \Ng memory.

From the results in \Cref{ablation}, we can observe that all components improve the performance of our model. Specifically, $\mathcal{L}_{\text{mm}}$ yields about  
{\textbf{1.51}$\%$ } performance boost as it brings samples of the same relation closer to each other while enforcing larger distances among different relation distributions. The $\mathcal{L}_{\text{pm}}$ improves the accuracy by 
{\textbf{3.18}$\%$}, which demonstrates the  effect of contrasting with the nearest wrong label. The adoption of $\mathcal{L}_{\text{con}}$ leads to 
{\textbf{1.28}$\%$} improvement, which shows that generating {hard negative} samples for memory data can help to better remember previous relational knowledge. {To better investigate the influence of $\mathcal{L}_{con}$, we conduct experiments with different $\lambda_{con}$ and show the results in \Cref{con_ablation}. We can see that the model achieves the best accuracy of 53.38 with $\lambda_{con}=0.02$ while the accuracy is only 52.13 with $\lambda_{con} =0.5$. In addition, the performance of the variant without $\mathcal{L}_{con}$ is worse than the performance of all other variants, which demonstrates the effectiveness of $\mathcal{L}_{con}$.}

The data augmentation module improves the performance by 
{\textbf{1.72}$\%$} as it can extract informative samples from unlabeled text which provide more relational knowledge for few-shot tasks. The results of variants without entity matching or similarity search verify that the two data augmentation methods are generally complementary to each other.

One could argue that the data augmentation module increases the complexity of \daer\ compared to other models. However, astute readers can find that even without data augmentation, \daer\  outperforms IDLVQ-C significantly for all tasks (compare `\daer\ \emph{w.o.} DA' with the baselines in \Cref{5shot}).

\paragraph{\daer's Performance under \crl}

\begin{figure}[t]
    \centering
    \includegraphics[width=0.46\textwidth]{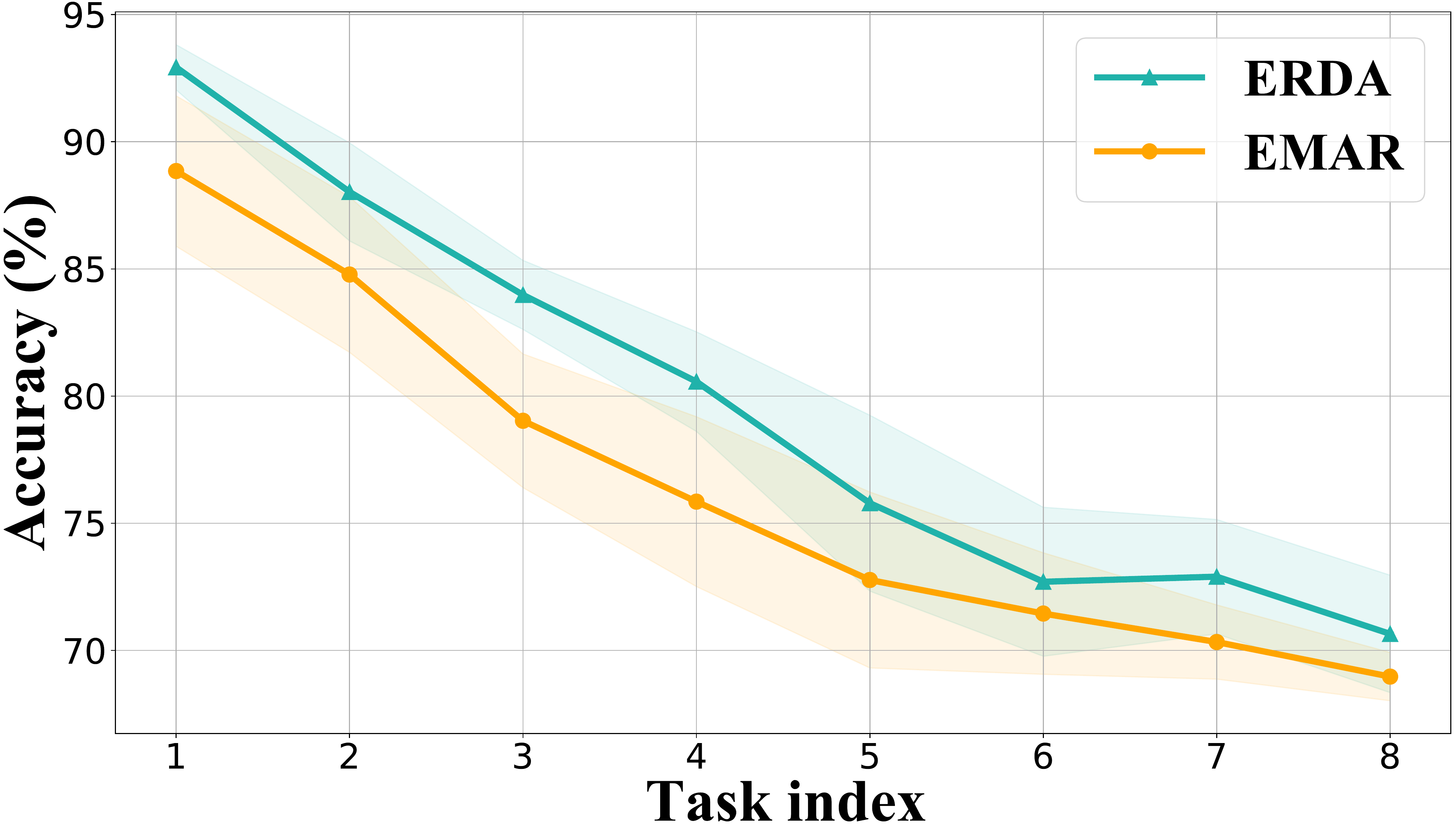}
    \label{Fig.crl}
    \caption{\label{crl}
    \small {Relation extraction results for ERDA (our) and EMAR \citep{han-etal-2020-continual} on the FewRel benchmark under the {\crl} setting. We randomly split the 80 relations into 8 tasks, where each task contains 10 relations. And we sample 100 examples per relation. From this figure, we can observe that ERDA outperforms EMAR in all \crl\ tasks.}
    }
\end{figure}

Although {\daer} is designed for \crfl, we also evaluate the embedding space regularization {(`\daer\ \emph{w.o.} DA')} in the {\crl} setting. {We sample 100 examples per relation for every task in FewRel and compare our method with the state-of-the-art method EMAR. The results are shown in \Cref{crl}. We can see that {\daer} outperforms EMAR in all tasks by \textbf{1.25} - \textbf{4.95}$\%$ proving that the embedding regularization can be a general method for \crl.}

\begin{table}[t!] \small
\setlength{\tabcolsep}{1.5pt}
\centering
\resizebox{1.00\linewidth}{!}{%
\begin{tabular}{llcccccccc}
\toprule
\multicolumn{2}{l}{\multirow{2}*{\textbf{Method}}}& \multicolumn{8}{c}{\textbf{Task index}}\\
\cmidrule{3-10}
\multicolumn{2}{c}{} & 1 & 2 & 3 & 4 & 5 & 6 & 7 & 8 \\
\midrule
\multicolumn{2}{l}{\daer}        & $\mathbf{92.57}$ & $\mathbf{79.17}$ & $\mathbf{70.43}$ & $\mathbf{65.01}$ & $\mathbf{61.06}$ & $\mathbf{57.54}$ & $\mathbf{54.88}$ & $\mathbf{53.23}$\\
\multicolumn{2}{l}{\emph{w.o.} $\mathcal{L}_{\text{mm}}$}    & $91.67$ & $78.38$ & $70.21$ & $63.77$ & $60.23$ & $56.32$ & $53.45$ & $51.72$ \\
\multicolumn{2}{l}{\emph{w.o.} $\mathcal{L}_{\text{pm}}$}       & $91.37$ & $75.80$ & $67.11$ & $61.13$ & $57.14$ & $54.04$ & $51.59$ & $50.05$ \\
\multicolumn{2}{l}{\emph{w.o.} $\mathcal{L}_{\text{con}}$}             & $91.63$ & $79.05$ & $69.28$ & $63.86$ & $59.66$ & $56.68$ & $54.12$ & $51.95$ \\
\multicolumn{2}{l}{\emph{w.o.} DA}            & $92.57$ & $77.84$ & $69.76$ & $63.74$ & $58.31$ & $56.12$ & $53.21$ & $51.51$ \\
\multicolumn{2}{l}{{\emph{w.o.} EM}}            & $92.57$ & $78.33$ & $70.17$ & $64.18$ & $59.63$ & $57.10$ & $54.18$ & $52.39$ \\
\multicolumn{2}{l}{{\emph{w.o.} SS}}            & $92.57$ & $78.56$ & $69.94$ & $63.98$ & $59.85$ & $56.92$ & $53.75$ & $52.27$ \\
\multicolumn{2}{l}{\emph{w.o.} M}    & $91.95$ & $77.59$ & $66.47$ & $57.08$ & $51.08$ & $47.36$ & $43.88$ & $40.32$\\
\bottomrule
\end{tabular}
}
\caption{\label{ablation}
\small Ablations on \textbf{FewRel} benchmark (\textbf{$10$-way $5$-shot}). The variance over 6 runs is reported in \Cref{ablation-with-var} in Appendix. We show the analysis of `\emph{w.o.} M' in Appendix ~\ref{sec:memory}.
}
\end{table}

%% file: Sections/appendix.tex
\section{Appendix}
\label{sec:appendix}


\subsection{Block Diagram of \daer\ Training} \label{sec:appendix2}

\begin{figure}[h]
    \centering
    \includegraphics[width=0.24\textwidth]{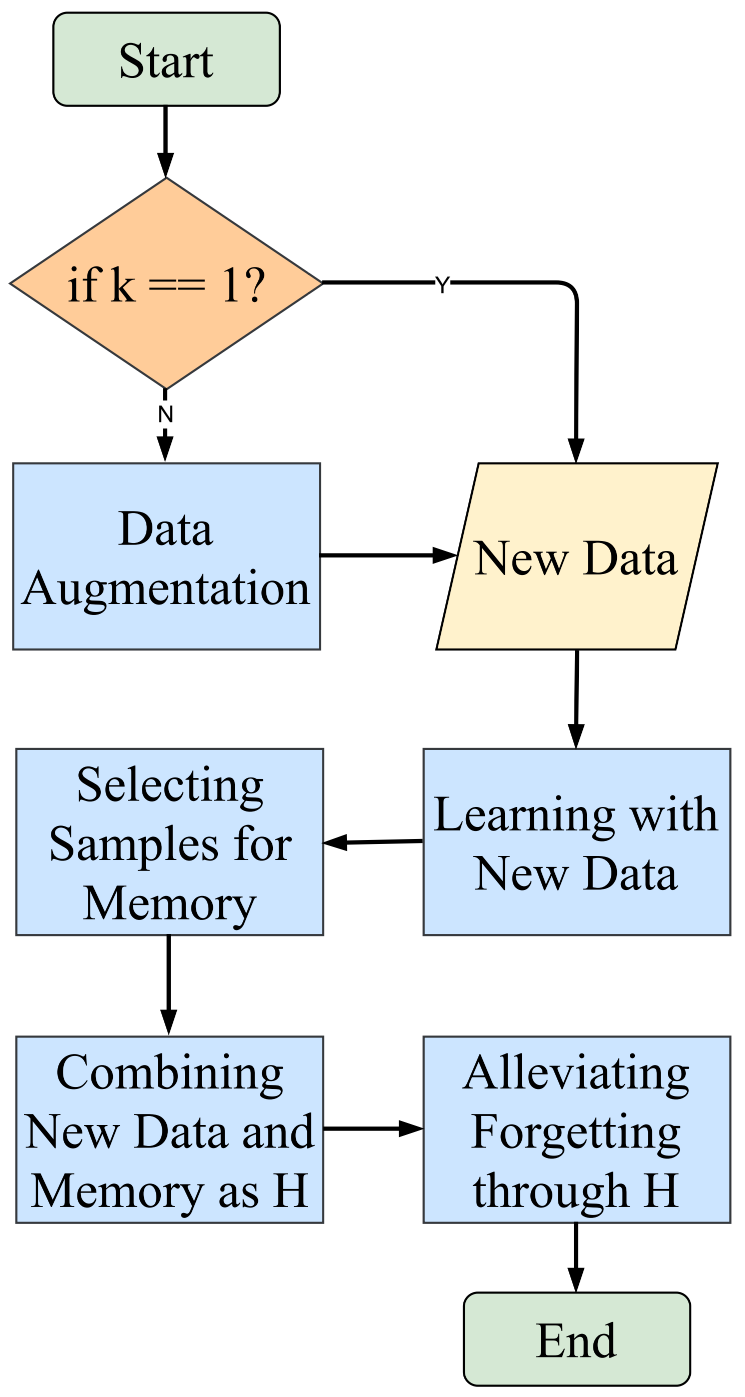}
    \label{Fig.diagram}
    \caption{\label{diagram}
    \small The block diagram of \daer's training at time step $k$.
    }
\end{figure}

\subsection{Hyperparameter Search} \label{sec:appendix3}
We follow the settings in \newcite{han-etal-2020-continual} for the Bi-LSTM encoder to have a fair comparison. For data augmentation, we set the threshold $\alpha=0.65$ and the number of samples selected by Faiss ($K$) as 1. We adopt $0.2, 0.2$ and $0.01$ for the three margin values $m_1, m_2$ and $m_3$, respectively. The loss weights $\lambda_{\text{ce}}, \lambda_{\text{mm}}, \lambda_{\text{pm}}$ and $\lambda_{\text{con}}$ are set to 1.0, 1.0, 1.0 and 0.1, respectively. In \cref{alg:Framwork}, we set $1$ for ${iter}_1$ and $2$ for ${iter}_2$. Hyperparameter search 
is done on the validation sets. {We follow EMAR \citep{han-etal-2020-continual} and use a grid search to select the hyperparameters. Specifically, the search spaces are:}

\begin{itemize}[leftmargin=*]
    \item Search range for $\alpha$ is $\left[0.3, 0.8\right]$ with a step size of $0.05$. 
    \item Search range for $K$ is $\left[1, 3\right]$ with a step size of $1$.

    \item Search range for $m_1$ and $m_2$ is $\left[0.1, 0.3\right]$ with a step size of $0.1$.

    \item Search range for $m_3$ is $\left[0.01, 0.03\right]$ with a step size of $0.01$.
    
    \item Search range for ${iter}_2$ in \cref{alg:Framwork} is $\left[1, 3\right]$ with a step size of $1$.

\end{itemize}





\begin{table*}[ht!]
\centering
\resizebox{1.00\linewidth}{!}{%
\begin{tabular}{llcccccccc}
\toprule
\multicolumn{2}{l}{\multirow{2}*{\textbf{Method}}}& \multicolumn{8}{c}{\textbf{Task index}}\\
\cmidrule{3-10}
\multicolumn{2}{c}{} & 1 & 2 & 3 & 4 & 5 & 6 & 7 & 8 \\
\midrule
\multicolumn{2}{l}{SeqRun}          & $96.35_{\pm 0.25}$ & $70.23_{\pm 2.42}$ & $58.13_{\pm 2.08}$ & $54.17_{\pm 1.90}$ & $48.82_{\pm 3.42}$ & $43.52_{\pm 2.45}$ & $37.90_{\pm 1.93}$ & $33.97_{\pm 1.53}$\\
\multicolumn{2}{l}{Joint Training}  & $96.35_{\pm 0.25}$ & $87.85_{\pm 2.25}$ & $82.87_{\pm 2.69}$ & $\mathbf{80.05}_{\pm 2.61}$ & $\mathbf{77.62}_{\pm 1.89}$ & $\mathbf{74.69}_{\pm 1.04}$ & $\mathbf{72.23}_{\pm 0.68}$ & $\mathbf{69.74}_{\pm 0.34}$\\
\midrule
\multicolumn{2}{l}{EMR}             & $96.35_{\pm 0.25}$ & $88.02_{\pm 2.09}$ & $78.83_{\pm 2.80}$ & $75.15_{\pm 2.85}$ & $72.00_{\pm 2.23}$ & $69.41_{\pm 2.06}$ & $66.70_{\pm 1.57}$ & $63.68_{\pm 1.47}$ \\
\multicolumn{2}{l}{EMAR}            & $92.03_{\pm 1.98}$ & $78.87_{\pm 3.72}$ & $72.81_{\pm 5.25}$ & $69.19_{\pm 4.45}$ & $68.05_{\pm 4.08}$ & $66.23_{\pm 1.95}$ & $63.68_{\pm 2.55}$ & $61.77_{\pm 1.48}$\\
\multicolumn{2}{l}{IDLVQ-C}         & $96.03_{\pm 0.12}$ & $87.18_{\pm 2.51}$ & $76.63_{\pm 3.97}$ & $73.57_{\pm 4.43}$ & $67.74_{\pm 3.60}$ & $65.16_{\pm 2.96}$ & $62.64_{\pm 1.87}$ & $60.32_{\pm 1.75}$ \\
\midrule
\multicolumn{2}{l}{\textbf{\daer}}   & $\mathbf{96.38}_{\pm 0.35}$ & $\mathbf{88.91}_{\pm 1.96}$ & $\mathbf{83.10}_{\pm 1.80}$ & $\mathbf{79.73}_{\pm 2.69}$ & $\mathbf{74.83}_{\pm 3.06}$ & $\mathbf{72.84}_{\pm 1.75}$ & $\mathbf{70.28}_{\pm 1.79}$ & $\mathbf{68.07}_{\pm 1.94}$ \\
\bottomrule
\end{tabular}
}
\caption{
\small Accuracy ($\%$) of different methods with a BERT encoder on \textbf{FewRel} benchmark for \textbf{$10$-way $5$-shot} setting. \daer\ is significantly better than EMR with $p$-value $= 0.003$. 
}
\label{bert-5shot}
\end{table*}

\subsection{The Influence of Task Order} \label{sec:task_order}


To evaluate the influence of the task order, we show the results (ERDA and IDLVQ-C) of six different runs with different task order on the FewRel benchmark for \textit{10-way 5-shot} setting in \Cref{allruns}. 
From the results, we can see that the order of tasks will influence the performance. For example, ERDA achieves 55.59 accuracy after learning task8 on the second run while the accuracy after learning task8 on the fifth run is only 51.35. More importantly, ERDA outperforms IDLVQ-C by a large margin in all six different runs.

\begin{table*}[ht!]
\centering
\resizebox{0.75\linewidth}{!}{%
\begin{tabular}{llcccccccc}
\toprule
\multicolumn{2}{c}{\multirow{2}*{\textbf{Run index}}}& \multicolumn{8}{c}{\textbf{Task index}}\\
\cmidrule{3-10}
\multicolumn{2}{c}{} & 1 & 2 & 3 & 4 & 5 & 6 & 7 & 8 \\
\midrule
\multicolumn{2}{c}{\multirow{2}*{{1}}} & \textbf{93.42} & \textbf{77.60} & \textbf{68.13} & \textbf{65.77} & \textbf{62.66} & \textbf{59.72} & \textbf{52.09} & \textbf{54.39} \\
\multicolumn{2}{c}{}  & 91.40 & 65.30 & 50.00 & 49.23 & 50.28 & 46.22 & 41.64 & 42.96 \\
\midrule
\multicolumn{2}{c}{\multirow{2}*{{2}}} & 91.02 & \textbf{76.55} & \textbf{68.03} & \textbf{62.32} & \textbf{57.26} & \textbf{54.73} & \textbf{56.97} & \textbf{55.59} \\
\multicolumn{2}{c}{}  & \textbf{92.10} & 61.10 & 49.37 & 44.88 & 40.90 & 43.72 & 42.43 & 38.47 \\
\midrule
\multicolumn{2}{c}{\multirow{2}*{{3}}} & \textbf{93.32} & \textbf{81.30} & \textbf{74.37} & \textbf{68.77} & \textbf{66.00} & \textbf{58.47} & \textbf{55.70} & \textbf{52.76} \\
\multicolumn{2}{c}{}  & 92.30 & 76.40 & 66.70 & 52.11 & 50.12 & 45.92 & 42.16 & 39.64 \\
\midrule
\multicolumn{2}{c}{\multirow{2}*{{4}}} & \textbf{92.42} & \textbf{77.50} & \textbf{64.50} & \textbf{57.90} & \textbf{60.12} & \textbf{52.87} & \textbf{52.53} & \textbf{53.65} \\
\multicolumn{2}{c}{}  & 92.20 & 62.65 & 57.30 & 51.73 & 51.26 & 46.00 & 42.81 & 40.04 \\
\midrule
\multicolumn{2}{c}{\multirow{2}*{{5}}} & \textbf{93.02} & \textbf{82.10} & \textbf{73.83} & \textbf{66.60} & \textbf{59.98} & \textbf{60.78} & \textbf{56.09} & \textbf{51.35} \\
\multicolumn{2}{c}{}  & 92.30 & 72.45 & 60.47 & 51.25 & 46.82 & 45.27 & 39.19 & 38.36 \\
\midrule
\multicolumn{2}{c}{\multirow{2}*{{6}}} & 92.22 & \textbf{80.00} & \textbf{73.73} & \textbf{68.70} & \textbf{60.36} & \textbf{58.67} & \textbf{55.90} & \textbf{51.64} \\
\multicolumn{2}{c}{}  & \textbf{93.10} & 77.00 & 60.67 & 60.75 & 56.48 & 50.32 & 45.26 & 43.89 \\
\bottomrule
\end{tabular}
}
\caption{
\small Accuracy ($\%$) of six different runs with different task order on \textbf{FewRel} benchmark for \textbf{$10$-way $5$-shot} setting. For every run, the upper row is the result of ERDA and the lower row shows the performance of IDLVQ-C.}
\label{allruns}
\end{table*}




\subsection{The Contribution of Memory} \label{sec:memory}
We conduct the ablation without memory (`\emph{w.o.} M') to analyze the contribution of the memory module on the FewRel \textit{10-way 5-shot} setting. From the results in \Cref{ablation-with-var}, we can observe that ERDA shows much better performance than `\emph{w.o.} M', which verifies the importance of the memory module. In addition, comparing the results of `\emph{w.o.} M' and `SeqRun' in \Cref{5shot}, we can find that `\emph{w.o.} M' achieves much better accuracy. This demonstrates the effectiveness of improving the representation ability of the model through margin-based losses.

\begin{table*}[ht!]
\centering
\resizebox{1.00\linewidth}{!}{%
\begin{tabular}{llcccccccc}
\toprule
\multicolumn{2}{l}{\multirow{2}*{\textbf{Method}}}& \multicolumn{8}{c}{\textbf{Task index}}\\
\cmidrule{3-10}
\multicolumn{2}{c}{} & 1 & 2 & 3 & 4 & 5 & 6 & 7 & 8 \\
\midrule
\multicolumn{2}{l}{SeqRun}          & $92.78_{\pm 0.76}$ & $52.11_{\pm 2.06}$ & $30.08_{\pm 1.75}$ & $24.33_{\pm 2.38}$ & $19.83_{\pm 0.99}$ & $16.90_{\pm 0.99}$ & $14.36_{\pm 0.69}$ & $12.34_{\pm 0.61}$\\
\multicolumn{2}{l}{Joint Training}  & $\mathbf{92.78}_{\pm 0.76}$ & $76.29_{\pm 3.47}$ & $69.39_{\pm 3.18}$ & $64.75_{\pm 2.48}$ & $60.45_{\pm 1.67}$ & $\mathbf{57.64}_{\pm 0.84}$ & $52.80_{\pm 0.99}$ & $50.03_{\pm 1.17}$\\
\midrule
\multicolumn{2}{l}{EMR}             & $92.78_{\pm 0.76}$ & $69.14_{\pm 2.74}$ & $56.24_{\pm 3.32}$ & $50.03_{\pm 2.91}$ & $46.50_{\pm 2.30}$ & $43.21_{\pm 1.47}$ & $39.88_{\pm 1.25}$ & $37.51_{\pm 1.53}$ \\
\multicolumn{2}{l}{EMAR}            & $85.20_{\pm 4.15}$ & $62.02_{\pm 3.34}$ & $52.45_{\pm 3.75}$ & $48.95_{\pm 5.46}$ & $46.77_{\pm 2.56}$ & $44.33_{\pm 2.83}$ & $40.75_{\pm 2.60}$ & $39.04_{\pm 2.05}$\\
\multicolumn{2}{l}{IDLVQ-C}         & $92.23_{\pm 0.50}$ & $69.15_{\pm 6.42}$ & $57.42_{\pm 6.14}$ & $51.66_{\pm 4.74}$ & $49.31_{\pm 4.72}$ & $46.24_{\pm 2.00}$ & $42.25_{\pm 1.79}$ & $40.56_{\pm 2.13}$ \\
\midrule
\multicolumn{2}{l}{\textbf{\daer}}   & $92.57_{\pm 0.82}$ & $\mathbf{79.17}_{\pm 2.08}$ & $\mathbf{70.43}_{\pm 3.75}$ & $\mathbf{65.01}_{\pm 3.84}$ & $\mathbf{61.06}_{\pm 2.71}$ & $\mathbf{57.54}_{\pm 2.80}$ & $\mathbf{54.88}_{\pm 1.86}$ & $\mathbf{53.23}_{\pm 1.49}$ \\
\bottomrule
\end{tabular}
}
\caption{
\small Accuracy ($\%$) and variance of different methods at every time step on \textbf{FewRel} benchmark for \textbf{$10$-way $5$-shot} \crfl.
}
\label{5shot-with-var}
\end{table*}

\begin{table*}[ht!]
\centering
\resizebox{1.00\linewidth}{!}{%
\begin{tabular}{llcccccccc}
\toprule
\multicolumn{2}{l}{\multirow{2}*{\textbf{Method}}}& \multicolumn{8}{c}{\textbf{Task index}}\\
\cmidrule{3-10}
\multicolumn{2}{c}{} & 1 & 2 & 3 & 4 & 5 & 6 & 7 & 8 \\
\midrule
\multicolumn{2}{l}{\daer}        & $\mathbf{92.57}_{\pm 0.82}$ & $\mathbf{79.17}_{\pm 2.08}$ & $\mathbf{70.43}_{\pm 3.75}$ & $\mathbf{65.01}_{\pm 3.84}$ & $\mathbf{61.06}_{\pm 2.71}$ & $\mathbf{57.54}_{\pm 2.80}$ & $\mathbf{54.88}_{\pm 1.86}$ & $\mathbf{53.23}_{\pm 1.49}$\\
\multicolumn{2}{l}{\emph{w.o.} $\mathcal{L}_{\text{mm}}$}    & $91.67_{\pm 1.00}$ & $78.38_{\pm 2.70}$ & $70.21_{\pm 4.23}$ & $63.77_{\pm 4.03}$ & $60.23_{\pm 2.78}$ & $56.32_{\pm 3.13}$ & $53.45_{\pm 2.11}$ & $51.72_{\pm 1.27}$ \\
\multicolumn{2}{l}{\emph{w.o.} $\mathcal{L}_{\text{pm}}$}       & $91.37_{\pm 0.60}$ & $75.80_{\pm 3.82}$ & $67.11_{\pm 4.63}$ & $61.13_{\pm 2.47}$ & $57.14_{\pm 2.81}$ & $54.04_{\pm 2.36}$ & $51.59_{\pm 2.30}$ & $50.05_{\pm 1.14}$ \\
\multicolumn{2}{l}{\emph{w.o.} $\mathcal{L}_{\text{con}}$}             & $91.63_{\pm 0.64}$ & $79.05_{\pm 2.46}$ & $69.28_{\pm 1.95}$ & $63.86_{\pm 2.77}$ & $59.66_{\pm 3.14}$ & $56.68_{\pm 2.55}$ & $54.12_{\pm 1.18}$ & $51.95_{\pm 1.15}$ \\
\multicolumn{2}{l}{\emph{w.o.} DA}            & $92.57_{\pm 0.82}$ & $77.84_{\pm 4.07}$ & $69.76_{\pm 2.62}$ & $63.74_{\pm 3.89}$ & $58.31_{\pm 2.38}$ & $56.12_{\pm 2.97}$ & $53.21_{\pm 2.32}$ & $51.51_{\pm 0.70}$ \\
\multicolumn{2}{l}{{\emph{w.o.} EM}}            & $92.57_{\pm 0.82}$ & $78.33_{\pm 2.73}$ & $70.17_{\pm 4.34}$ & $64.18_{\pm 2.82}$ & $59.63_{\pm 2.22}$ & $57.10_{\pm 1.73}$ & $54.18_{\pm 1.79}$ & $52.39_{\pm 0.66}$ \\
\multicolumn{2}{l}{{\emph{w.o.} SS}}            & $92.57_{\pm 0.82}$ & $78.56_{\pm 3.64}$ & $69.94_{\pm 3.04}$ & $63.98_{\pm 2.56}$ & $59.85_{\pm 2.18}$ & $56.92_{\pm 2.56}$ & $53.75_{\pm 2.05}$ & $52.27_{\pm 0.98}$ \\
\multicolumn{2}{l}{\emph{w.o.} M}    & $91.95_{\pm 0.82}$ & $77.59_{\pm 2.28}$ & $66.47_{\pm 2.04}$ & $57.08_{\pm 3.08}$ & $51.08_{\pm 2.60}$ & $47.36_{\pm 4.88}$ & $43.88_{\pm 1.29}$ & $40.32_{\pm 2.22}$\\
\bottomrule
\end{tabular}
}
\caption{\label{ablation-with-var}
\small Accuracy ($\%$) and variance of the ablations on \textbf{FewRel} benchmark (\textbf{$10$-way $5$-shot}).
}
\end{table*}